\documentclass[conference]{IEEEtran}
\IEEEoverridecommandlockouts

\usepackage{cite}
\usepackage{amsmath,amssymb,amsfonts}
\usepackage{algorithmic}
\usepackage{graphicx}
\usepackage{flushend}
\usepackage[capitalise]{cleveref}
\usepackage{textcomp}
\usepackage{xcolor,colortbl}
\usepackage[top=54pt, left=54pt, right=54pt, bottom=54pt]{geometry}%
\def\BibTeX{{\rm B\kern-.05em{\sc i\kern-.025em b}\kern-.08em
    T\kern-.1667em\lower.2ex\hbox{E}\kern-.125emX}}

\definecolor{Gray}{gray}{0.75}
\definecolor{Grey}{gray}{0.92}

\newcolumntype{a}{>{\columncolor{Grey}}c}

\title{Exploring the Potential of Multi-modal Sensing Framework for Forest Ecology}

\begin{document}
\date{}

\author{
Luca Romanello$^{1,2}$, Tian Lan$^{1}$, Mirko Kovac$^{2,3}$, Sophie F. Armanini$^{1}$, Basaran Bahadir Kocer$^{2,4}$  \\

\thanks{ 
\par $^{1}$eAviation Laboratory, TUM School of Engineering and Design, Munich.
\par $^{2}$Aerial Robotics Laboratory, Imperial College London.
\par $^{3}$Laboratory of Sustainability Robotics, EMPA, Dübendorf, Switzerland.
\par $^{4}$School of Civil, Aerospace and Design Engineering, University of Bristol.}
}
\maketitle

\section{Extended Abstract}

Forests offer essential resources and services to humanity, yet preserving and restoring them presents challenges, particularly due to the limited availability of actionable data, especially in hard-to-reach areas like forest canopies. Accessibility continues to pose a challenge for biologists collecting data in forest environments, often requiring them to invest significant time and energy in climbing trees to place sensors. This operation not only consumes resources but also exposes them to danger.
Efforts in robotics have been directed towards accessing the tree canopy using robots. A swarm of drones has showcased autonomous navigation through the canopy, maneuvering with agility and evading tree collisions, all aimed at mapping the area and collecting data \cite{doi:10.1126/scirobotics.abm5954}. However, relying solely on free-flying drones has proven insufficient for data collection. Flying drones within the canopy generates loud noise, disturbing animals and potentially corrupting the data. Additionally, commercial drones often have limited autonomy for dexterous tasks where aerial physical interaction could be required, further complicating data acquisition efforts.
Aerial deployed sensor placement methods such as bio-gliders \cite{biodegradable_gliders, 20} and sensor shooting \cite{sensors_delivery} have proven effective for data collection within the lower canopy. However, these methods face challenges related to retrieving the data and sensors, often necessitating human intervention.

There are noteworthy attempts, with inspiration from nature, to use physical interaction within the forest ecosystem \cite{kocer2021forest}. Sloths, for example, hug and cling to branches as they provide natural protection from predators. By imitating this approach, \cite{zheng2023metamorphic,  doi:10.1089/soro.2022.0010,  10380767, askari2024crash} were capable of perching on branches saving energy opening the possibility to collect data without creating additional noise. 
While these initiatives represent significant progress and have inspired further exploration in the realm of physical interactions, they are not without limitations. Current frameworks can only perch on branches of limited size, and they require ample open space to maneuver within the canopy to reach different data collection points. Furthermore, the perching maneuver, executed from the bottom side of the vehicles, restricts their capabilities and demands complex control algorithms for execution. AVOCADO \cite{10323196} tries to tackle this last issue by using a tethered pod approach. The aim was to analyse the gradient in height of the tree with an unwinding system which allows it to go up and down after is perched on a tree. 

The challenges of the aforementioned studies rely on a single modality, which should be made manually with human intervention or by hovering with a drone while the pod is going up and down, limiting its potential. We took the initiative to improve the motion modality considering a tethered perching approach \cite{hauf2023learning} and proposed an effective system consisting of (i) a drone; (ii) a ring mechanism; (iii) a tether; and (iv) a suspended pod with a two-rotor system. With the propeller and tether winding system on the suspended pod, our system provided both (i) manoeuvre within the canopy; and (ii) perching and disentangling from tree branches to enable an idle time to collect long-term data. Our system implements a ring system to connect the tether, which allows the drone to rotate 360 around the tether so as to perch on the branch \cite{lan2024aerial}. This approach has no limit in branch size and shape, as the tether is inherently complaint to surfaces. We demonstrated perching with both the drone and the suspended pod, further improving the pod capability which is not merely connected to the sensing of the environment, but with a perching capability. We demonstrated perching and disentangling from branches which can be done by both drone and pod, with pod and tether winding being an energy-efficient way to do it.

\begin{figure}
    \centering
	\includegraphics[width=1\columnwidth]{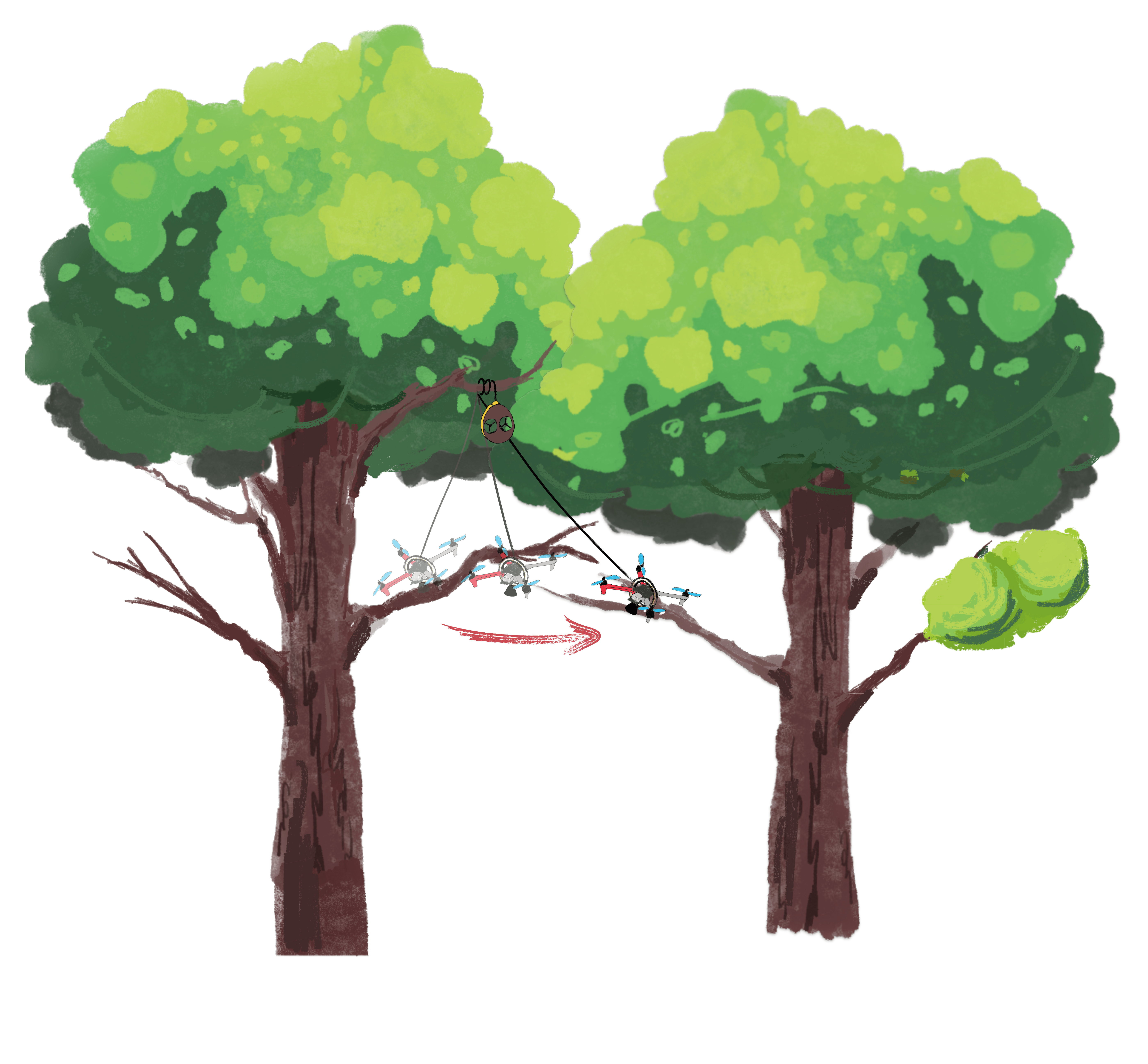}
    \label{pic_drone}
    \caption{The proposed system: a drone connected with a suspended pod demonstrating multi-modal motion in a perched state. The tether allows the system to move between the branches in energy energy-efficient manner. This also allows for the creating of less noise and the proposed multi-modal system can traverse dense forests by swinging the system followed by perching and disentangling.}
\end{figure}

The next generation of robots for forest sensing needs to be designed in order to avoid as much the creation of noise, to provide the robots with physical interaction capabilities to improve their energy efficiency and to actually make the robot cohabit with the environment. 
Our efforts in this context are now pointing towards exploring the multi-sensing capability of this framework. The drone, which so far has only been used as a vector for transporting the environmental payload represented by the pod, now is actively involved in the sensing mission. The ring brings the advantage of adding a degree of freedom to the drone, which now can move in a more agile manner. Moreover, thanks to the tether connection and the perching on a branch, in case of a collision with a tree branch, leaf or trunk when manoeuvring within the canopy, the mission would not completely fail as it would happen with a no-tethered drone. For example, with our proposed system, the mission could be restarted as the drone would swing around the perched branch. All these features allow the drone to move within the canopy and it can be taken advantage of to move to new branches and even other trees in proximity taking data not only on the perched tree but also on neighbour trees. Our proposed approach is illustrated in Fig. \ref{pic_drone}.

This framework can effectively be employed in a multi-sensing mission, where the pod is involved in a mission to scan and map the ecosystem along the gradient of the height of the tree, while the drone can move within the same height from one branch to another collecting other precious data. Moreover, the drone itself can be employed with a physical interaction capability such as the sloth drones, which would allow it to anchor on another branch, and save energy. Tests have been done to demonstrate the convenience of this approach also in the sense of energy consumption. Testing drones in some circular trajectories with and without tether, we demonstrate that with the ring and tether framework, the drone is capable of saving up to 8 times more energy. 

The reduction in energy consumption also has a positive impact on noise levels. As the drone requires less thrust to move, it consequently produces less noise, thereby preserving the ecosystem it operates within. We envision that such a system can allow for collecting long-term data including vision \cite{ho2022vision} and sound \cite{lawson2023use}.

\bibliographystyle{IEEEtran}
\bibliography{mybib}

\end{document}